\documentclass[10pt,twocolumn,letterpaper]{article}

\usepackage[pagenumbers]{iccv} 

\usepackage[ruled,linesnumbered]{algorithm2e}
\usepackage{setspace}
\usepackage{multirow}
\usepackage{diagbox}
\usepackage{colortbl}
\usepackage[dvipsnames]{xcolor}

\definecolor{iccvblue}{rgb}{0.21,0.49,0.74}
\usepackage[pagebackref,breaklinks,colorlinks,allcolors=iccvblue]{hyperref}

\def\paperID{8401} 
\def\confName{ICCV}
\def\confYear{2025}

\title{When Lighting Deceives: Exposing Vision-Language Models' Vulnerability through Illumination Transformation Attack}

\author{
Hanqing Liu$^{\dagger}$, Shouwei Ruan$^{\dagger}$, Yao Huang$^{\dagger}$, Shiji Zhao, Xingxing Wei\thanks{Corresponding author.} \\
Institute of Artificial Intelligence, Beihang University
}

\begin{document}
\twocolumn[{%
\renewcommand\twocolumn[1][]{#1}%
\maketitle

\begin{center}
\centering
\vspace{-3ex}
\includegraphics[height=9cm]{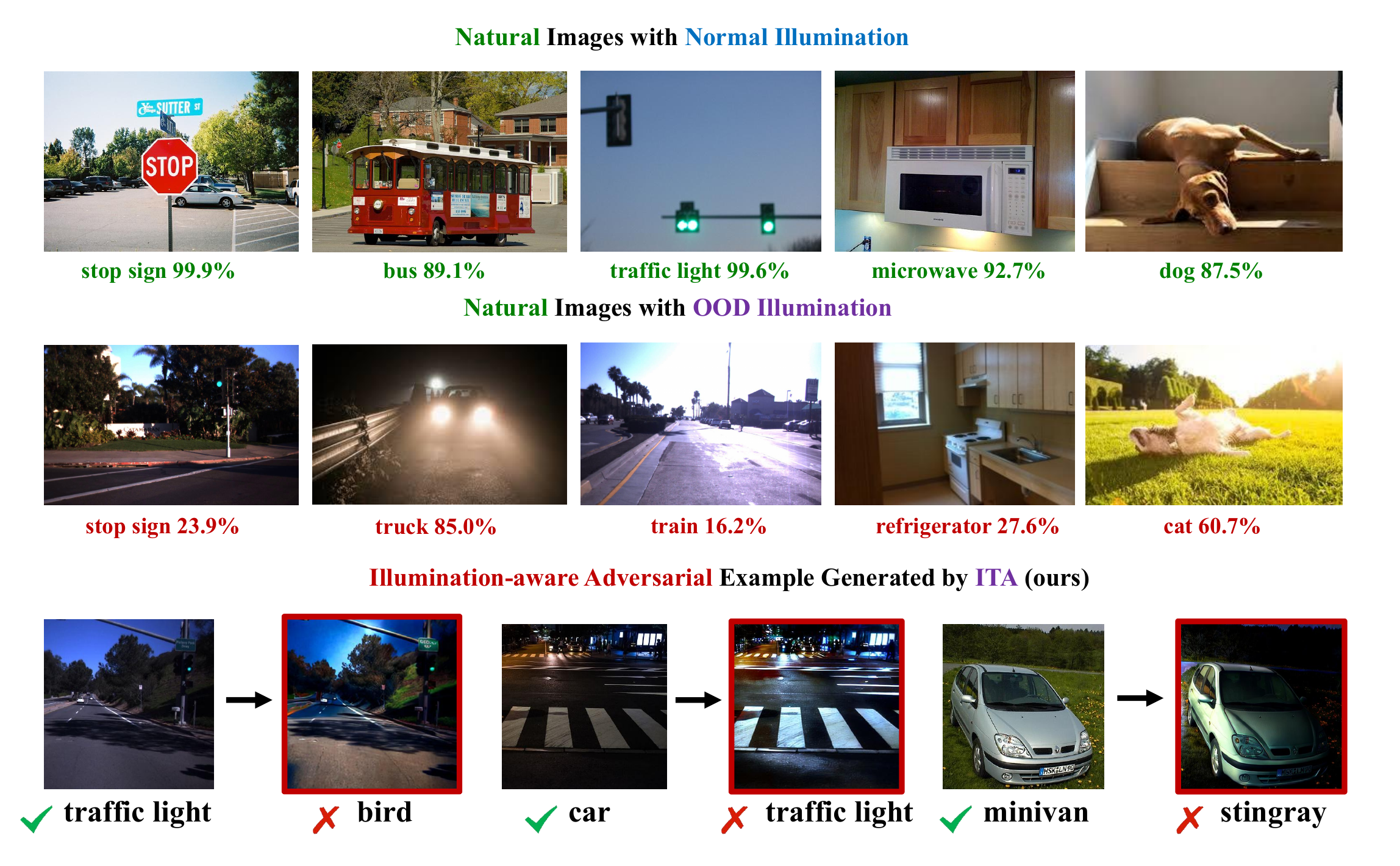}
\captionof{figure}{\textbf{Revealing VLM Vulnerabilities to Illumination Variations.} Comparison of the predictions made by OpenCLIP ViT-B/16 on natural images with normal and OOD illumination versus illumination-aware adversarial examples generated by our \textbf{Illumination Transformation Attack (ITA)}, which illustrate VLMs are vulnerable to illumination changes, underscoring the need for systematic evaluation.}
\vspace{-1ex}
\label{fig1}
\end{center}%
}]

\renewcommand{\thefootnote}{}
\footnotetext{$\dagger$ Equal contribution.}
\renewcommand{\thefootnote}{\arabic{footnote}}

\begin{abstract}
 Vision-Language Models (VLMs) have achieved remarkable success in various tasks, yet their robustness to real-world illumination variations remains largely unexplored. To bridge this gap, we propose \textbf{I}llumination \textbf{T}ransformation \textbf{A}ttack (\textbf{ITA}), the first framework to systematically assess VLMs' robustness against illumination changes. However, there still exist two key challenges: (1) how to model global illumination with fine-grained control to achieve diverse lighting conditions and (2) how to ensure adversarial effectiveness while maintaining naturalness. To address the first challenge, we innovatively decompose global illumination into multiple parameterized point light sources based on the illumination rendering equation. This design enables us to model more diverse lighting variations that previous methods could not capture. Then, by integrating these parameterized lighting variations with physics-based lighting reconstruction techniques, we could precisely render such light interactions in the original scenes, finally meeting the goal of fine-grained lighting control. For the second challenge, by controlling illumination through the lighting reconstrution model's latent space rather than direct pixel manipulation, we inherently preserve physical lighting priors. Furthermore, to prevent potential reconstruction artifacts, we design additional perceptual constraints for maintaining visual consistency with original images and diversity constraints for avoiding light source convergence. 
 Extensive experiments demonstrate that our ITA could significantly reduce the performance of advanced VLMs, e.g., LLaVA-1.6, while possessing competitive naturalness, exposing VLMS' critical illuminiation vulnerabilities.
\end{abstract}
\section{Introduction}
\label{sec:intro}
Recently, Vision-Language Models (VLMs) have achieved remarkable progress in various tasks, such as image captioning~\cite{alayrac2022flamingo,li2023blip} and visual question answering~\cite{liu2024visual,liu2024llavanext,chen2024internvl}. Due to their superior performance, these models are increasingly being deployed in real-world applications like  autonomous driving~\cite{tian2024drivevlm,xu2024drivegpt4} and security surveillance~\cite{abba2024real, qu2024double}. However, in such real-world applications, they may inevitably face challenging environmental variations~\cite{al2024unibench, ruan2023improving, ruan2023towards, duan2021adversarial}. Among them, illumination changes are the most ubiquitous and unavoidable one. As shown in Fig.~\ref{fig1}, while VLMs perform well under normal illumination, they often misclassify images with  Out-Of-Distribution (OOD) illumination. These illumination shifts can dramatically alter scene perception without being immediately noticeable, posing significant security risks. Thus, it is urgent to prioritize and systematically investigate VLMs' illumination robustness across varying environmental conditions.

Existing research has made some efforts in exploring illumination impacts, particularly for specific tasks like traffic sign recognition~\cite{duan2021adversarial, zhong2022shadows, gnanasambandam2021optical, hsiao2024natural, sun2024embodied} and face recognition~\cite{nguyen2020adversarial, li2023physical}. They simulate illumination variations by generating localized light disturbances like spots, reflections, or laser interference. Nevertheless, they are primarily designed for CNN-based models and show significant limitations when applied to VLMs. The methods either employ overly simplistic illumination modifications (static lighting changes or shadow effects) or generate artifacts easily identifiable by VLMs. Moreover, current illumination attacks are lack of diversity, failing to capture the rich spectrum of real-world lighting conditions and variations.

Based on the above discussions, this paper aims to generate more diverse and natural illumination-aware adversarial examples for comprehensively evaluating VLMs' illumination robustness. However, there still exist two key challenges that must be addressed: \textbf{(1)} \textit{How to model global illumination with fine-grained control to achieve diverse lighting conditions that reflect real-world scenarios?} \textbf{(2)} \textit{How to ensure adversarial effectiveness while maintaining physical realism and natural appearance?}

To address these challenges, we propose a novel framework named \textbf{\underline{I}}llumination \textbf{\underline{T}}ransformation \textbf{\underline{A}}ttack (\textbf{ITA}). The ITA consists of two primary components: First, to achieve fine-grained illumination modeling, we innovatively decompose global illumination into multiple parameterized point light sources following the illumination rendering equation~\cite{kajiya1986rendering}. This approach allows us to express the illumination as a linear sum of contributions from individual light sources rather than a single light spot, enabling us to model diverse lighting variations that previous methods could not capture. Then, by leveraging physics-based lighting reconstruction techniques IC-Light~\cite{iclight}, we can accurately reconstruct illumination examples corresponding to specific illumination variations, ultimately achieving the intention of fine-grained lighting control.

Second, our approach prioritizes maintaining natural appearance in the generated images. By controlling illumination through IC-Light's latent space rather than direct pixel manipulation, we inherently preserve physical lighting priors and avoid unnatural lighting that violates physical constraints. Moreover, to prevent uncontrollable effects from the diffusion model's randomness, such as overexposure and underexposure, we implement perceptual constraints that maintain visual consistency with original images, alongside diversity constraints that prevent multiple light sources from converging. This paradigm ensures the generated illumination variations remain physically plausible while effectively revealing vulnerabilities in VLMs. 

In addition, IC-Light's non-differentiable nature resulting from diffusion models and denoising operations may prevent gradient-based optimization for adversarial attacks. Thus, to overcome this limitation, we employ the Covariance Matrix Adaptation Evolution Strategy (CMA-ES)~\cite{nomura2024cmaes}, a gradient-free evolutionary optimization algorithm, to iteratively optimize the illumination variation.
Finally, through extensive experiments, we demonstrate that illumination-aware adversarial examples generated by ITA significantly degrade the performance of advanced VLMs under optimized illumination conditions, revealing previously unidentified  vulnerabilities. The contributions of this paper can be summarized as follows:
\begin{itemize}
\item We propose a novel framework called ITA for generating illumination-aware adversarial examples, marking the first work to explore the illumination robustness of VLMs. It transcends previous methods' limitations, enabling more comprehensive illumination variations.

\item We innovatively formulate the generation of complex illumination scenarios as a parameterized optimization problem, enabling precise control over lighting conditions while generating illumination-aware adversarial examples that are both natural and deceptive.

\item Experimental results demonstrate that our illumination-aware adversarial examples successfully expose illumination sensitivity across various VLMs, including CLIP and its variants, as well as more complex generative VLMs such as LLaVA-1.6, BLIP-2, and Instruct-BLIP.

\end{itemize}
\section{Related Work}
\label{sec:Related Work}

\subsection{Vision-Language Models}
Owing to large-scale pretraining, 
Vision-Language Models (VLMs)~\cite{liu2024visual,liu2024llavanext,lu2019vilbert,li2022blip} have emerged as powerful tools for processing multimodal information, demonstrating exceptional capabilities in tasks requiring joint understanding of visual and textual inputs, e.g., image captioning, visual question answering (VQA), and complex reasoning tasks that bridge visual perception with language comprehension.
Despite their impressive achievements~\cite{radford2021learning,liu2025mmbench,zhang2024multitrust}, concerns remain regarding their robustness, particularly in challenging edge cases and safety-critical applications. Recent studies~\cite{tong2024eyes, ruan2024omniview} reveal that VLMs still struggle with fine-grained visual attributes, such as spatial orientation, numerical understanding, and color perception, which may stem from inherent limitations in their visual representations. Additionally, VLMs are vulnerable to adversarial perturbations in the $L_p$-norm space~\cite{cui2024robustness,mao2022understanding}.

While previous works have explored risks caused by challenging visual attributes or adversarial perturbation, there remains a critical gap in evaluating VLMs' robustness under natural visual attribute like illumination variations, which is a fundamental aspect of real-world perception. To bridge it, our ITA will systematically examine how changes in illumination conditions impact VLM performance.

\subsection{Illumination Adversarial Attacks}
Existing research on illumination-based adversarial attacks has explored several approaches to evaluate visual model's robustness. Early methods focused on direct light manipulation, including projected light patterns~\cite{gnanasambandam2021optical, huang2022spaa}, laser beam interference~\cite{duan2021adversarial}, naturalistic adversarial patches~\cite{hu2021naturalistic}, and shadow manipulations~\cite{zhong2022shadows}. These optical attacks, however, face significant limitations in bright environments where environmental illumination overwhelms the manipulated light signals. More recent approaches have attempted to address these limitations. Wang et al.~\cite{wang2023rfla} developed reflection-based adversarial illumination strategies, while Hsiao et al.~\cite{hsiao2024natural} introduced natural light attack methods that maintain effectiveness under stronger ambient lighting conditions. Despite these advances, current illumination attack research remains constrained in several ways: (1) evaluations have primarily targeted specific applications like traffic sign recognition; (2) most studies focus on conventional CNN-based vision models rather than modern Vision-Language Models (VLMs), and (3) existing methods typically optimize for single light sources, failing to capture the real-world illumination variations.

In contrast, our proposed ITA framework extends adversarial illumination attacks to broader and more practical scenarios. Through simultaneously optimizing multiple light sources, our approach significantly broadens the spectrum of illumination variations, enabling a comprehensive evaluation of VLMs' robustness to illumination conditions. Moreover, our approach ensures high perceptual realism, making the adversarial examples more natural and indistinguishable from real-world illumination conditions.

\section{Methodology}
\label{sec:Methodology}

\begin{figure*}[t]
  \centering
  \includegraphics[width=\textwidth]{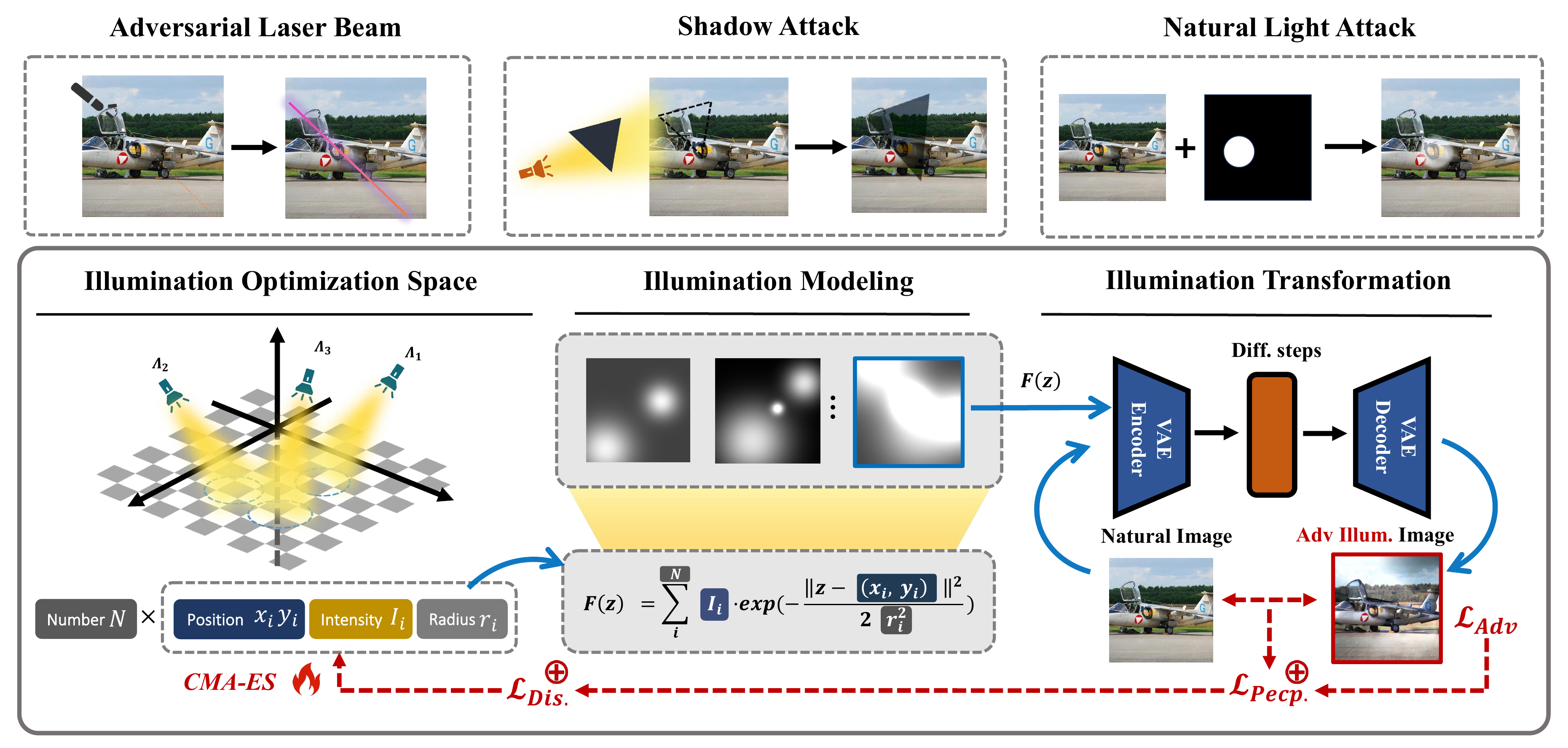}
  \vspace{-2ex}
  \caption{\textbf{Overview of the Proposed ITA Framework.} Three previously proposed illumination attack methods rely on localized lighting disturbances~\textbf{(top)}. In contrast, our method applies illumination transformations across the entire scene, generating illumination-aware adversarial examples, ensuring both naturalness and adversariality~\textbf{(bottom)}.}
  \vspace{-2ex}
    \label{fig:framework}
\end{figure*}

In this section, we introduce the proposed ITA framework, as presented in Fig.~\ref{fig:framework}. The ITA framework will use the IC-Light technique to generate illumination transformation samples with given illumination variations. Thus, we begin by providing background on IC-Light, followed by a discussion of illumination modeling and optimization strategy.

\subsection{Preliminary} 
\label{Preliminary}
Given an input image \( X \), the conditional inputs consist of an optional text input \( T \) and an optional light source input \( X_l \). The vanilla IC-Light model employs a variational auto encoder (VAE) $\mathcal{E}_e$ and a CLIP text encoder $E_t$ to map the inputs into a latent space, which can be formulated as:
\begin{equation}
    z_i = \mathcal{E}_e(X, X_l), \quad z_t = E_t(T). 
\end{equation}

The encoded representations are then concatenated and passed as cross-attention inputs to the UNet. The UNet undergoes a diffusion process and generates a transformed output, which is decoded back into pixel space using the decoder $\mathcal{E}_d$, producing illumination variations in the image, which can be simply expressed as:
\begin{equation}
    {X'} = \mathcal{E}_d(\text{Diff}(z_i, z_t)),
\end{equation}
where \( X' \) denotes the generated illumination transformation samples, and \(\text{Diff}(\cdot) \) represents the diffusion process. However, while this method is effective for reconstructing illumination transformations, it lacks fine-grained control over localized illumination conditions and is non-differentiable due to the inclusion of diffusion models. To address this, we will model the illumination distribution in Sec.~\ref{Illumination Modeling} and convert it into an optimization problem in Sec.~\ref{Illumination Optimization}.

\subsection{Illumination Modeling}
\label{Illumination Modeling}
To achieve precise control over illumination conditions, we innovatively model the scene lighting using multiple point light sources. Each light source is characterized by four key parameters: spatial position $(x, y)$, intensity $I$, and radius $r$ of influence. The complete lighting configuration $\boldsymbol{\Lambda}$ represents our parameterization of the illumination environment:
\begin{equation}
\boldsymbol{\Lambda} = \{ (x_i, y_i, I_i, r_i) \}_{i=1}^{N},
\label{problem1}
\end{equation}
where $N$ is the number of light sources. Then, to simulate realistic light behavior, we apply a Gaussian function to model the spatial distribution of each light source:
\begin{equation}
F_i(z) = I_i \cdot \exp\left(-\frac{\| z - (x_i, y_i) \|^2}{2r_i^2}\right),
\label{problem2}
\end{equation}
where $z$ represents any point in the scene, with $F_i(z)$ quantifying the illumination contribution from the $i$-th light source at that point. The parameters $(x_i, y_i)$, $I_i$, and $r_i$ from $\boldsymbol{\Lambda}$ directly control the center position, brightness, and spread of each light source respectively.

The total illumination at any point is calculated as the sum of all individual light contributions:
\begin{equation}
F(z) = \sum_{i=1}^{N} F_i(z).
\end{equation}
This formulation enables us to generate diverse and realistic lighting effects by adjusting the parameters within $\boldsymbol{\Lambda}$.

Above all, the benefit of our illumination modeling is twofold. First, unlike traditional approaches that struggle with optimizing complex nonlinear light distributions, our method preserves essential physical light properties while making the parameters easier to optimize. Second, we address limitations of previous illumination attack methods that typically use binary masks (1 for illuminated regions, 0 for non-illuminated areas)  to simulate lighting effects locally, which often lead to visual discontinuities. Our Gaussian-based model holistically reconstructs illumination across the entire scene rather than applying isolated modifications, eliminating the artifacts and inconsistencies common in localized lighting modifications. 
\subsection{Optimization Objective of ITA Framework}
\label{Illumination Optimization}
\noindent\textbf{Adversarial Objective.} The adversarial objective of our illumination transformation attack (ITA) is to learn an optimal adversarial illumination configuration \( \boldsymbol{\Lambda}^* \) that generates illumination conditions capable of significantly degrading the performance of VLMs, which can be formulated as:
\begin{equation}
    \boldsymbol{\Lambda}^* = \arg \max_{\boldsymbol{\Lambda}} \mathcal{L}_{Adv}( \mathcal{I}(X, \boldsymbol{\Lambda}),Y),
\label{problem3}
\end{equation}
where \( \mathcal{I}(X, \boldsymbol{\Lambda}) \) denotes the relighting process applying the illumination configuration \( \boldsymbol{\Lambda} \) to the original image \( X \), \( Y \) represents the truth label and \( \mathcal{L}_{Adv}(\cdot) \) is designed to quantify the effectiveness of the adversarial process.

In this work, we conduct all experiments with different versions of CLIP as the target model, leveraging its image encoder and text encoder to guide the adversarial optimization. Specifically, our adversarial attack objective is inspired by traditional adversarial attacks on classification models, where the goal is to reduce the predicted probability of the correct class. Given that CLIP can be viewed as an implicit classifier operating in an embedding space, we formulate our objective in a similar manner. To sum up, we minimize the probability assigned by the model to the ground-truth semantic label, effectively pushing the generated adversarial image \(X'\) towards a misclassification.

Given an adversarial image \(X'\) and a semantic label set \(Y = \{ y_i \}_{i=1}^{N}\) containing the ground-truth label \(y_t\), we first compute a semantic similarity vector as follows:
\begin{equation}
V = \{ cos \left[ E_v(X'),E_t(Y_i) \right] \}_{i=1}^{N},
\end{equation}
where \(cos(\cdot)\) denotes cosine similarity, \(E_v(\cdot)\) and \(E_t(\cdot)\) represent the image encoder and text encoder, respectively. The vector $V$ captures the alignment between \(X'\) and each semantic label. We then compute the probability distribution over semantic labels using the softmax function. Finally, the adversarial loss \(\mathcal{L}_{Adv}\) is defined as the negative log-likelihood of the ground-truth semantic label:
\begin{equation}
\mathcal{L}_{Adv}(X', Y) = -log (\frac{exp(V_t)}{{\textstyle \sum_{j=1}^{N}} exp(V_j)}).
\end{equation}

This objective directly reduces the confidence of CLIP in assigning the correct label, making it more likely to misclassify \(X'\). At the same time, we need to ensure that the generated images do not suffer from extreme illumination variations that lead to overexposure or underexposure. To address this, we propose a perceptual constraint. Additionally, when optimizing with multiple light sources, we introduce a distance regularization loss as a diversity constraint to prevent the converging of light sources that are too close to each other. The complete optimization further incorporates these constraints to maintain visual realism and encourage diverse light placements.

\noindent\textbf{Regularization for Perception and Diversity.} To ensure the perceptual naturalness of illumination adversarial examples, we impose constraints on both perception and diversity. For the perceptual constraint, we use the LPIPS (Learned Perceptual Image Patch Similarity) loss~\cite{zhang2018unreasonable}, which compares deep features from a pre-trained network to ensure that the adversarial image remains visually similar to the original. LPIPS aligns more closely with human perception than traditional metrics like L2 or SSIM, preserving structural details and texture consistency. This minimizes distortions, improving the stealthiness of the attack and making it harder for VLMs to distinguish adversarial examples from natural images, which can be formulated as:
\begin{equation}
    \mathcal{L}_{Pecp}(X, X') = \sum_{l} \left\| f_l(X) - f_l(X') \right\|_2^2,
\end{equation}
where \( f_l(\cdot) \) denotes the feature of the image at layer \( l \) of the pre-trained network.

\begin{algorithm}[!h] 
\small
\setstretch{0.3}
\caption{\small Optimization Algorithm}\label{algorithm}
\KwData{Original images $X$, Semantic label list $Y$, Number of light sources $N$, Initial distribution parameters $\boldsymbol{\mu}_0, \boldsymbol{C}_0, \boldsymbol{\Sigma}_0$, Weight coefficients $\alpha$, $\beta$.}
\KwResult{Adversarial illumination configuration $\boldsymbol{\Lambda}^*$, Adversarial illumination samples $X'$.}
\tcc{\scriptsize Image-Text Pairs Preparation}
$\{X_0,Y_0 \} \leftarrow \{X,Y \}$\;
\tcc{\scriptsize Initialization of distribution parameters}
$\boldsymbol{\mu}_0 \leftarrow \mathbf{0}$, $\boldsymbol{C_0} (\Sigma_0)^2 \leftarrow \mathbf{I}$\;
\While{$t < t_{\max}$}{
    \tcc{\scriptsize Sampling: Generate adversarial illumination parameters}
    \For{$i = 1 \to K$}{
        $\mathbf{q}_i^{t+1} \sim \mathcal{N}(\boldsymbol{\mu}_t, \boldsymbol{C_t} (\Sigma_t)^2)$\;
        $\boldsymbol{\Lambda}_i^{t+1} \leftarrow \mathbf{A} \cdot \tanh(\mathbf{q}_i^{t+1}) + \mathbf{B}$\;
        $(X')_i^{t+1} \leftarrow \mathcal{I}(X, \boldsymbol{\Lambda}_i^{t+1})$\;
        \tcc{\scriptsize Calculate loss value} 
        $\mathcal{L}_i^{t+1} = \mathcal{L}_{Adv}((X')_i^{t+1}, Y) + \alpha \cdot \mathcal{L}_{Pecp} + \beta \cdot \mathcal{L}_{Dis}$\;
        
        \If{should\_stop()}{
            \text{break}
        }
    }
    \tcc{\scriptsize Evaluation: Compute the objective function values for each sample}
    Sorting $\mathcal{L}_i^{t+1}$ in ascending order\;
    
    \tcc{\scriptsize Update: Distribution parameters update}
    $\boldsymbol{\mu}_{t+1} \leftarrow \sum_{i=1}^{K} w_i \mathbf{q}_i^{t+1}$\;
    $\boldsymbol{C}_{t+1} \leftarrow (1 - c_c) \boldsymbol{C}_t + c_c \sum_{i=1}^{K} w_i (\mathbf{q}_i^{t+1} - \boldsymbol{\mu}_t)(\mathbf{q}_i^{t+1} - \boldsymbol{\mu}_t)^T$\;
    $\boldsymbol{\Sigma}_{t+1} \leftarrow \boldsymbol{\Sigma}_t \cdot \exp\left( c_\sigma \left( \frac{\|\mathbf{p}\|}{\mathbb{E}[\|\mathcal{N}(0, I)\|]} - 1 \right) \right)$\;
}
$\boldsymbol{\mu}^* \leftarrow \boldsymbol{\mu}_{t_{\max}},
\boldsymbol{C}^* \leftarrow \boldsymbol{C}_{t_{\max}}, \boldsymbol{\Sigma}^* \leftarrow \boldsymbol{\Sigma}_{t_{\max}}$\;
$\mathbf{q}^* \sim \mathcal{N}(\boldsymbol{\mu}^*, \boldsymbol{C^*} (\Sigma^*)^2)$\;
$\boldsymbol{\Lambda}^* \leftarrow \mathbf{A} \cdot \tanh(\mathbf{q}^*) + \mathbf{B}$, $X' \leftarrow \mathcal{I}(X, \boldsymbol{\Lambda}^*)$\;

\end{algorithm}

To further enhance the effectiveness and diversity of the adversarial attack, we introduce a distance regularization loss to address the converging of multiple light sources. Without proper constraints, the optimization process may collapse into a degenerate solution, where all light sources converge to a single point, reducing the attack to a simple single-light perturbation. To ensure sufficient spatial separation between light sources and maintain the adversarial effectiveness, we define the distance regularization loss as:
\begin{equation}
    \mathcal{L}_{Dis} = \sum_{i=1}^{N} \sum_{j=i+1}^{N} \max(0, \boldsymbol{\delta} - \| \boldsymbol{\Lambda}_i-\boldsymbol{\Lambda}_j \| ),
\end{equation}
where $\boldsymbol{\delta}$ is a predefined minimum distance threshold. This penalty term encourages spatial separation between light sources, promoting a more diverse and effective adversarial lighting configuration. Finally, the total loss function combines the adversarial loss, the perceptual loss, and the distance loss with weighting factors \(\alpha\) and \(\beta\) as follow:
\begin{equation}
    \mathcal{L}_{Total} = \mathcal{L}_{Adv} + \alpha \cdot \mathcal{L}_{Pecp} + \beta \cdot \mathcal{L}_{Dis}.
\end{equation}

\subsection{Optimization Algorithm} 
\label{Algorithm}
In this section, we present our illumination optimization algorithm, which aims to obtain the best illumination configuration \( \boldsymbol{\Lambda}^* \). Rather than optimizing for a single configuration, we explore a distribution of configurations, which offers several key advantages: \textbf{1)} It enables a more diverse exploration of possible results, helping to identify regions of the configuration space where optimal illumination conditions may exist. \textbf{2)} It also facilitates faster convergence to a region containing the best illumination configuration, reducing the risk of getting trapped in local optima and enhancing robustness by avoiding overfitting to a solution, thus ensuring better generalization across different VLMs.

Specifically, our method optimizes adversarial illumination distributions using the Covariance Matrix Adaptation Evolution Strategy (CMA-ES)~\cite{hansen2016cma,golovin2017google}, a gradient-free optimization algorithm well-suited for this task. Since it does not rely on gradient information, CMA-ES is ideal for optimizing adversarial illumination distributions. To further enhance the optimization process, we incorporate two techniques: Learning Rate Adaptation (LRA) and Early Stopping Policy. LRA dynamically adjusts the learning rate during the search to improve convergence speed, while Early Stopping Policy halts the optimization when further improvements are unlikely, thus preventing unnecessary iterations and saving computational resources.

The adversarial illumination configuration is parameterized by a Gaussian variable \(\mathbf{q}\), which follows a normal distribution with mean \(\boldsymbol{\mu}\), step-size \(\boldsymbol{\Sigma}\), and covariance matrix \(\boldsymbol{C}\). The optimization aims to improve the adversarial illumination conditions by adjusting the parameters \(\boldsymbol{\mu}\), \(\boldsymbol{C}\), and \(\boldsymbol{\Sigma}\). The optimization process proceeds through three main steps: \textbf{sampling}, \textbf{evaluation}, and \textbf{update}. For a detailed description of the overall process and mathematical formulations, which is refer to Algorithm~\ref{algorithm}. Additional information can be found in Appendix~\textcolor{iccvblue}{A}.

\section{Experiments}
\label{sec:Experiments}

\begin{table*}[!t]
\caption{\textbf{Performance Degradation (\textcolor{Maroon}{$\downarrow$}) of VLMs on Zero-shot Classification Under Illumination-aware Adversarial Examples Generated from COCO Dataset.} We compare the accuracy and natural scores evaluated by GPT-4o of previous methods with our proposed approach. The performance for ACC is considered better with lower values, while higher values are preferred for GPT-4o. Number in \textbf{bold} indicates the best performance and the second-best result is \underline{underlined}.}
\setlength\tabcolsep{4.8pt}
\renewcommand\arraystretch{1.0}
\centering
\scalebox{0.85}{
\begin{tabular}{c|cc|cc|cc|cc}
\hline
\multirow{2}{*}{\textbf{Method}} &
  \multicolumn{2}{c|}{\textbf{OpenCLIP ViT-B/16}} &
  \multicolumn{2}{c|}{\textbf{Meta-CLIP ViT-L/14}} &
  \multicolumn{2}{c|}{\textbf{EVA-CLIP ViT-G/14}} &
  \multicolumn{2}{c}{\textbf{OpenAI CLIP ViT-L/14}} \\ \cline{2-9} 
                     & ACC(\%)  & GPT-4o & ACC(\%)  & GPT-4o & ACC(\%)  & GPT-4o & ACC(\%) & GPT-4o \\ \hline

Clean                & 97                   & 3.65                          & 98                    & 3.65                          & 98                    & 3.65                          & 93                    & 3.65      \\ \hline
Natural Light Attack~\cite{hsiao2024natural} & 94$(\textcolor{Maroon}{\downarrow\! 3})$                    & \underline{2.32}$(\textcolor{Maroon}{\downarrow\! 1.33})$                         & 97$(\textcolor{Maroon}{\downarrow\! 1})$                    & 2.14$(\textcolor{Maroon}{\downarrow\! 1.51})$                         & 97$(\textcolor{Maroon}{\downarrow\! 1})$                    & 2.16$(\textcolor{Maroon}{\downarrow\! 1.49})$                         & 93(-)                    & 2.07$(\textcolor{Maroon}{\downarrow\! 1.58})$                         \\ \hline
Shadow Attack~\cite{zhong2022shadows}        & 84$(\textcolor{Maroon}{\downarrow\! 13})$                    & 2.11$(\textcolor{Maroon}{\downarrow\! 1.54})$                         & 82$(\textcolor{Maroon}{\downarrow\! 16})$                    & 1.91$(\textcolor{Maroon}{\downarrow\! 1.74})$                         & 95$(\textcolor{Maroon}{\downarrow\! 3})$                    & 2.03$(\textcolor{Maroon}{\downarrow\! 1.62})$                         & 79$(\textcolor{Maroon}{\downarrow\! 14})$                    & 1.78$(\textcolor{Maroon}{\downarrow\! 1.87})$                         \\ \hline
ITA(random)        & \underline{71}$(\textcolor{Maroon}{\downarrow\! 26})$                    & \textbf{2.56}$(\textcolor{Maroon}{\downarrow\! 1.09})$                         & \underline{81}$(\textcolor{Maroon}{\downarrow\! 17})$                    & \textbf{2.24}$(\textcolor{Maroon}{\downarrow\! 1.41})$                         & \underline{93}$(\textcolor{Maroon}{\downarrow\! 5})$                    & \textbf{2.48}$(\textcolor{Maroon}{\downarrow\! 1.17})$                         & \underline{67}$(\textcolor{Maroon}{\downarrow\! 26})$                    & \textbf{2.33}$(\textcolor{Maroon}{\downarrow\! 1.32})$                         \\ \hline
ITA(best)          & \textbf{46$(\textcolor{Maroon}{\downarrow\! 51})$}           & 2.11$(\textcolor{Maroon}{\downarrow\! 1.54})$                         & \textbf{64$(\textcolor{Maroon}{\downarrow\! 34})$}           & \underline{2.15}$(\textcolor{Maroon}{\downarrow\! 1.50})$                         & \textbf{84$(\textcolor{Maroon}{\downarrow\! 14})$}           & \underline{2.17}$(\textcolor{Maroon}{\downarrow\! 1.48})$                         & \textbf{51$(\textcolor{Maroon}{\downarrow\! 42})$}           & \underline{2.15}$(\textcolor{Maroon}{\downarrow\! 1.50})$                         \\ \hline
\end{tabular}}
\label{exp:1}
\end{table*}

\begin{figure*}[!h]
  \centering
  \includegraphics[width=\textwidth]{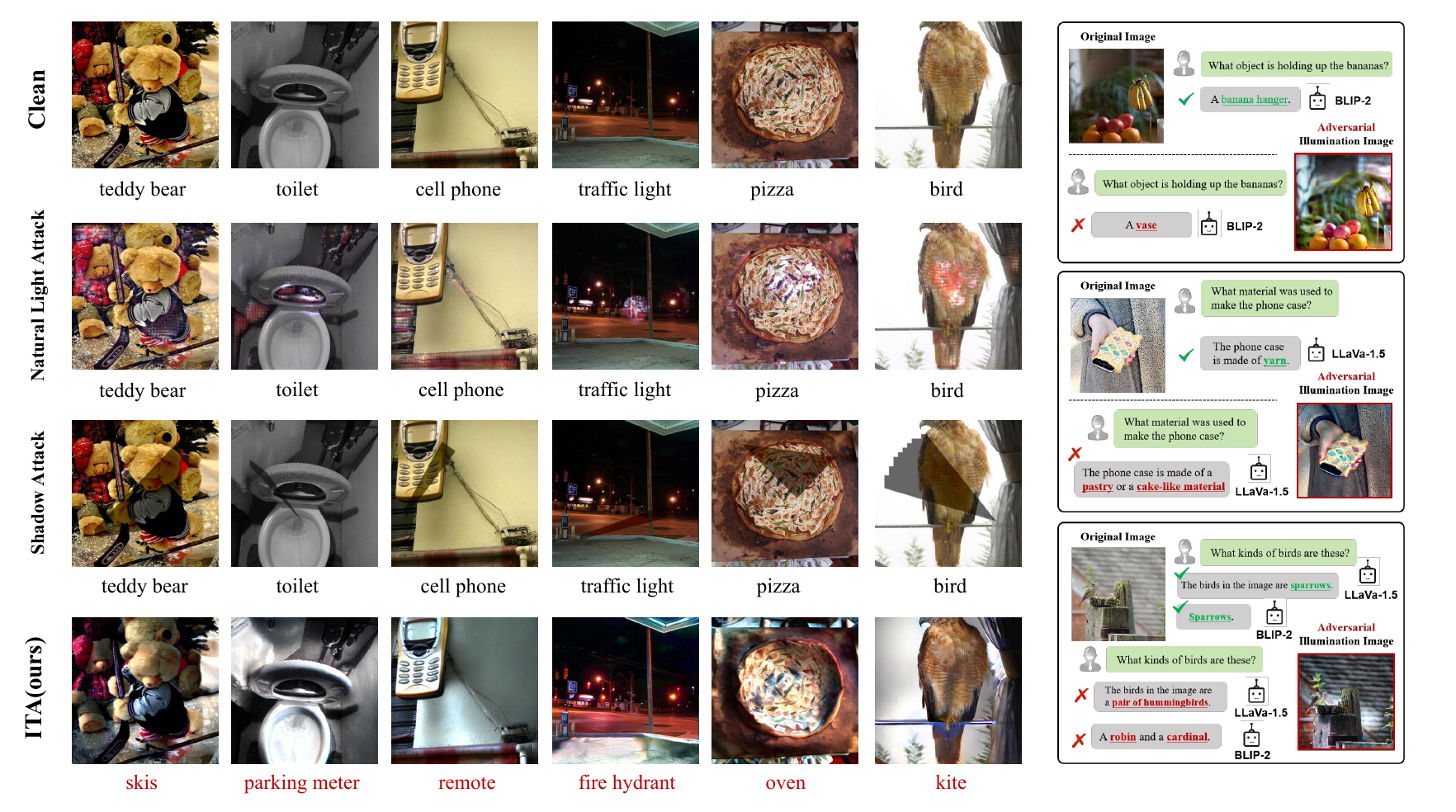}
  \caption{\textbf{Visualization Results.} From up to down, the images represent the results of Clean, Natural Light Attack, Shadow Attack, and our method, along with their corresponding Top-1 labels, evaluated on OpenCLIP ViT-B/16. The red color indicates \textcolor{Maroon}{misclassified labels}.\textbf{(Left)} Visualization of some illumination changes causing LVLMs to give incorrect answers.\textbf{(Right)}}
  \label{fig:vis1}
\end{figure*}

In this section, we demonstrate the effectiveness of our proposed method \textbf{ITA} for crafting adversarial illuminating transformation examples against VLMs.
\subsection{Experimental Settings}

\noindent\textbf{Datasets.} We evaluate ITA on three representative vision-language tasks \textbf{\emph{(zero-shot image classification, image captioning, Visual Question Answering)}} using only the COCO dataset~\cite{chen2015microsoft}. This approach ensures more accurate evaluations, as the same dataset provides real labels for all three tasks, eliminating the potential influence of dataset distribution. Additionally, we carefully select 30 categories, totaling 300 images with clear instance from COCO. This consistency across tasks allows for a more reliable and integrated assessment of ITA’s performance. We use COCO-80 categories as our semantic label list, and the selected categories are provided in Appendix.~\textcolor{iccvblue}{B}.

\noindent\textbf{Threat Models.} For the image classification task, we select the foundation models, a diverse set of CLIP variants, including OpenCLIP ViT-B/16~\cite{cherti2023reproducible,openclip}, Meta-CLIP ViT-L/14~\cite{xu2023demystifying}, EVA-CLIP ViT-G/14~\cite{sun2023eva} and OpenAI CLIP ViT-L/14~\cite{radford2021learning}. This selection is motivated by the need to evaluate the robustness of our proposed methods across different versions and parameter scales of CLIP models. For image captioning and VQA tasks, we choose current mainstream VLM models, including LLaVA-1.5~\cite{liu2024visual}, LLaVA-1.6~\cite{liu2024improved}, OpenFlamingo~\cite{awadalla2023openflamingo}, Blip-2 ViT-L~\cite{li2023blip}, Blip-2 FlanT5-XL~\cite{li2023blip} and InstructBLIP~\cite{liu2023visual}, covering a range of architectures and capabilities. The proposed method is compared to the following baselines: Shadows Attack\cite{zhong2022shadows} and Natual Light Attack\cite{hsiao2024natural}. We follow the same setup as reported in the original work, with the only difference being the substitution of the target model for the attack.

\noindent\textbf{Evaluation Metrics.} For the image classification task, we evaluate top-1 accuracy and report the relative drop compared to clean samples. We employ a 4-point scale to assess the naturalness of the images using GPT-4o, focusing primarily on three fields: Visual Naturalness, Physical Consistency, and Adversarial Likelihood. Additionally, for image captioning and VQA tasks, we adopt the LLM-as-a-judge evaluation framework, following recent studies~\cite{liu2024visual,zheng2023judging,zhang2024multitrust}. Specifically, we design two distinct templates to assess the consistence scores of generated captions and the correctness of responses to given questions. The details of these evaluation settings are provided in Appendix.~\textcolor{iccvblue}{C}.

\noindent\textbf{Implementation Details.} In our experiments, we set the number of samples per iteration to 20 and run a total of 200 iterations. The number of light sources is 3 and the initial light distribution parameters are randomly assigned and the spatial boundaries \((x, y)\) range from 0 to the corresponding image dimensions, intensity \(I\) is set between 0.5 and 1.0, and radius \(r\) varies between 10 and 50. The distance threshold in distance penalty loss is set to 50. The weighting factors $\alpha$ and $\beta$ are 0.1 and 0.01 respectively. Additionally, we establish two baseline settings: the original natural images and randomly generated illumination transformation samples without optimization. All experiments are conducted on 8 NVIDIA RTX 4090 24GB GPUs, and the detailed computational cost is provided in Appendix.~\textcolor{iccvblue}{D}.

\subsection{Comparisons with Other Methods} \textbf{Evaluation on Zero-Shot Classification Tasks.} Tab.~\ref{exp:1} presents the comparison of our method with previous illumination attack methods on zero-shot classification tasks. Our method consistently outperforms others across all CLIP versions, achieving a notable \textbf{54\%} attack success rate on OpenCLIP ViT-B/16 while maintaining a high naturalness score. This highlights the robustness of our approach, which misleads the model effectively while preserving image quality. Additionally, the accuracy degradation across different models varies significantly. Smaller models, like OpenCLIP ViT-B/16, suffer the most (a 51\% drop), while larger models like EVA-CLIP ViT-G/14 and Meta-CLIP ViT-L/14 models show a more moderate drop in accuracy (around 14\% and 34\%, respectively). This suggests that models with more parameters might be slightly more resilient, but they still face considerable performance degradation under illumination transformation. We also observe a trade-off between image naturalness and classification accuracy. As the accuracy drops, the naturalness score generally decreases. Overall, our method balances adversariality with naturalness, outperforming previous techniques. As shown in Fig.~\ref{fig:vis1}, visualizations demonstrate that our adversarial examples appear more natural compared to others, with no alteration in the image's semantic content. Additional examples can be found in Appendix.~\textcolor{iccvblue}{E}.

\begin{table*}[!t]
\caption{\textbf{Performance Degradation (\textcolor{Maroon}{$\downarrow$}) of VLMs on Image Captioning Under Illumination-aware Adversarial Examples Generated from COCO Dataset.} We compare the consistency (\%) evaluated by GPT-4 of previous methods with our proposed approach. The performance is considered better with lower values. Number in \textbf{bold} indicates the best performance and the second-best result is \underline{underlined}.}
\renewcommand\arraystretch{1.0}
\centering
\scalebox{0.72}{
\begin{tabular}{l|cc|cccccc}
\hline
\textbf{Image Encoder} & \textbf{Models} & \textbf{\#Params} & \textbf{Clean} & \textbf{Natural Light Attack}~\cite{hsiao2024natural} & \textbf{Shadow Attack}~\cite{zhong2022shadows} & \textbf{ITA(random)} & \textbf{ITA(best)} \\ \hline
\multirow{4}{*}{OpenAI CLIP ViT-L/14} & LLaVA-1.5 & 7B     & 78.60 & 77.00$(\textcolor{Maroon}{\downarrow\! 1.63})$ & 74.60$(\textcolor{Maroon}{\downarrow\! 4.00})$ & \underline{69.04}$(\textcolor{Maroon}{\downarrow\! 9.56})$ & \textbf{63.73$(\textcolor{Maroon}{\downarrow\! 14.87})$} \\ \cline{2-8} 
                           & LLaVA-1.6 & 7B & 72.10 & 71.70$(\textcolor{Maroon}{\downarrow\! 0.39})$ & 71.17$(\textcolor{Maroon}{\downarrow\! 0.93})$ & \underline{69.19}$(\textcolor{Maroon}{\downarrow\! 2.91})$ & \textbf{61.60$(\textcolor{Maroon}{\downarrow\! 10.51})$} \\ \cline{2-8} 
                           & OpenFlamingo & 3B         & 70.20 & 69.53$(\textcolor{Maroon}{\downarrow\! 0.67})$ & 67.80$(\textcolor{Maroon}{\downarrow\! 2.40})$ & \underline{66.37}$(\textcolor{Maroon}{\downarrow\! 3.85})$ & \textbf{53.93$(\textcolor{Maroon}{\downarrow\! 16.27})$} \\ \cline{2-8} 
                           & Blip-2 (FlanT5$_\text{XL}$ ViT-L) & 3.4B & 75.10 & 70.77$(\textcolor{Maroon}{\downarrow\! 4.33})$ & 68.57$(\textcolor{Maroon}{\downarrow\! 6.53})$ & \underline{67.40}$(\textcolor{Maroon}{\downarrow\! 7.73})$ & \textbf{60.93$(\textcolor{Maroon}{\downarrow\! 14.17})$} \\ \hline
\multirow{2}{*}{EVA-CLIP ViT-G/14} & Blip-2 (FlanT5$_\text{XL}$)  & 4.1B  & 74.96 & 71.27$(\textcolor{Maroon}{\downarrow\! 3.69})$ & \underline{68.80}$(\textcolor{Maroon}{\downarrow\! 6.17})$ & 69.50$(\textcolor{Maroon}{\downarrow\! 5.47})$ & \textbf{62.01$(\textcolor{Maroon}{\downarrow\! 11.88})$} \\ \cline{2-8} 
                               & InstructBLIP (FlanT5$_\text{XL}$)  & 4.1B   & 76.50 & 72.07$(\textcolor{Maroon}{\downarrow\! 4.43})$ & 69.77$(\textcolor{Maroon}{\downarrow\! 6.73})$ & \underline{69.03}$(\textcolor{Maroon}{\downarrow\! 7.97})$ & \textbf{63.20$(\textcolor{Maroon}{\downarrow\! 13.30})$} \\ \hline
\end{tabular}}
\label{exp:2}
\end{table*}

\begin{table*}[!t]
\caption{\textbf{Performance Degradation (\textcolor{Maroon}{$\downarrow$}) of VLMs on VQA Task Under Illumination-aware Adversarial Examples Generated from COCO Dataset.} We compare the correctness (\%) evaluated by GPT-4 of previous methods with our proposed approach. The performance is considered better with lower values. Number in \textbf{bold} indicates the best performance and the second-best result is \underline{underlined}.}
\renewcommand\arraystretch{1.0}
\centering
\scalebox{0.72}{
\begin{tabular}{l|cc|cccccc}
\hline
\textbf{Image Encoder} & \textbf{Models} & \textbf{\#Params} & \textbf{Clean} & \textbf{Natural Light Attack}~\cite{hsiao2024natural} & \textbf{Shadow Attack}~\cite{zhong2022shadows} & \textbf{ITA(random)} & \textbf{ITA(best)} \\ \hline
\multirow{4}{*}{OpenAI CLIP ViT-L/14} & LLaVA-1.5 & 7B        & 68.00 & 68.00(-) & 67.00 $(\textcolor{Maroon}{\downarrow\! 1.00})$ & \underline{59.00} $(\textcolor{Maroon}{\downarrow\! 9.00})$ & \textbf{48.00} $(\textcolor{Maroon}{\downarrow\! 20.00})$ \\  \cline{2-8} 
                        &LLaVA-1.6 & 7B    & 64.00 & 63.00 $(\textcolor{Maroon}{\downarrow\! 1.00})$ & 64.00 (-) & \underline{50.00}  $(\textcolor{Maroon}{\downarrow\! 14.00})$ & \textbf{43.00} $(\textcolor{Maroon}{\downarrow\! 21.00})$ \\ \cline{2-8} 
                        &OpenFlamingo & 3B    & 45.00 & 39.00 $(\textcolor{Maroon}{\downarrow\! 6.00})$ & 44.00 $(\textcolor{Maroon}{\downarrow\! 1.00})$ & \underline{40.00}  $(\textcolor{Maroon}{\downarrow\! 5.00})$ & \textbf{19.00} $(\textcolor{Maroon}{\downarrow\! 26.00})$ \\  \cline{2-8} 
                        & Blip-2 (FlanT5$_\text{XL}$ ViT-L) & 3.4B & 63.00 & 58.00 $(\textcolor{Maroon}{\downarrow\! 5.00})$ & 50.00  $(\textcolor{Maroon}{\downarrow\! 13.00})$ & \underline{48.00} $(\textcolor{Maroon}{\downarrow\! 15.00})$ & \textbf{38.00} $(\textcolor{Maroon}{\downarrow\! 25.00})$ \\ \hline
\multirow{2}{*}{EVA-CLIP ViT-G/14} & Blip-2 (FlanT5$_\text{XL}$)  & 4.1B   & 54.00 & 53.00 $(\textcolor{Maroon}{\downarrow\! 1.00})$ & 54.00(-) & \underline{51.00} $(\textcolor{Maroon}{\downarrow\! 3.00})$ & \textbf{33.00} $(\textcolor{Maroon}{\downarrow\! 21.00})$ \\  \cline{2-8} 
                        & InstructBLIP (FlanT5$_\text{XL}$)  & 4.1B & 68.00 & 64.00 $(\textcolor{Maroon}{\downarrow\! 4.00})$ & 62.00 $(\textcolor{Maroon}{\downarrow\! 6.00})$ & \underline{57.00} $(\textcolor{Maroon}{\downarrow\! 11.00})$ & \textbf{44.00} $(\textcolor{Maroon}{\downarrow\! 24.00})$ \\ \hline
\end{tabular}}
\label{exp:3}
\end{table*}

\noindent\textbf{Evaluation on Image Captioning \& VQA Tasks.} In addition to classification tasks, we also evaluate the effectiveness of our method on Image Captioning and VQA tasks using state-of-the-art large-scale Vision-Language Models (LVLMs). To ensure a fair and meaningful comparison, we perform transfer attacks by utilizing the samples generated by image encoders from different VLMs on zero-shot classification tasks. A clear trend is that smaller models with fewer parameters experience larger performance drops. For example, in Image Captioning (Tab.~\ref{exp:2}), OpenFlamingo and Blip-2  show significant degradation, particularly under the ITA, with performance drops of 16\% and 14\%, respectively. Larger models like LLaVA-1.5 and LLaVA-1.6 exhibit more moderate drops, around 10\%, suggesting that larger models are slightly more robust to illumination transformation. Similarly, in the VQA task (Tab.~\ref{exp:3}), smaller models like OpenFlamingo and Blip-2 show more substantial drops (up to 26\% and 25\%, respectively) under ITA. In contrast, larger models like InstructBLIP experience a smaller drop of 24\%, indicating some resilience but still significant vulnerability. In summary, our method is the most effective across both tasks, achieving the largest performance reductions, demonstrating the transferability of our method in attacking LVLMs across different tasks.

\begin{figure}[t]
  \centering
  \includegraphics[height=2.9cm]{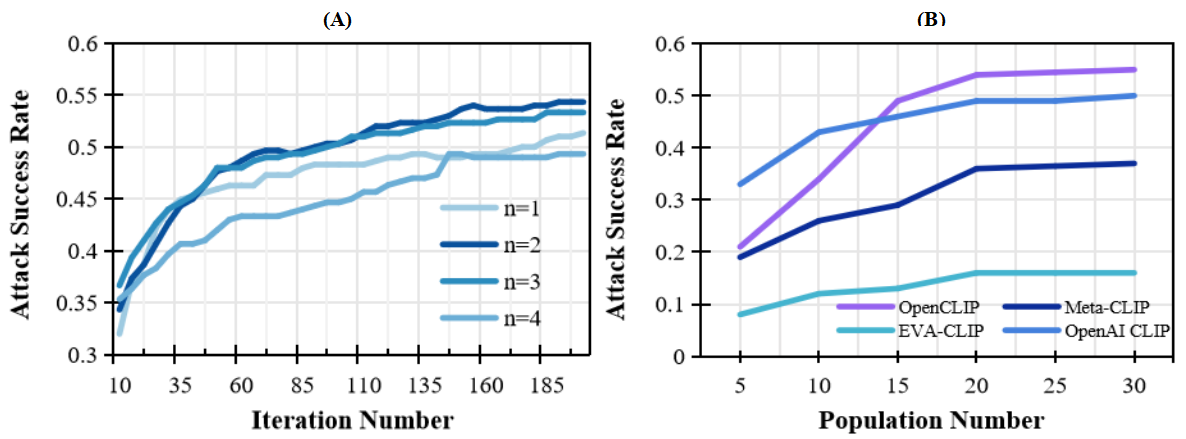}
    \caption{\textbf{Ablation Study Results on Optimization Hyperparameters.} Attack success rate of illumination-aware adversarial examples with (A) different numbers of light sources at different iteration steps on OpenCLIP and (B) different population number across different CLIP versions in 200 iteration steps.}
  \label{fig:xiaorong1}
\end{figure}

\subsection{Ablation Study}
\textbf{Optimization Hyperparameters.} As shown in Fig.~\ref{fig:xiaorong1}, we observe that the attack success rate continuously increases with the number of optimization iterations and converges after 200 steps. Based on this observation, we set the default value of the number of iterations to 200 throughout the entire experiment to balance effectiveness and computational efficiency. Additionally, we analyze the number of light sources optimized jointly. The results show that the highest attack success rate is achieved when \(n = 2\). Furthermore, we conduct experiments on the number of samples in each optimization iteration, and the results demonstrate that convergence is reached when the population size is 20.

\begin{figure}[t]
  \centering
  \includegraphics[height=3.4cm]{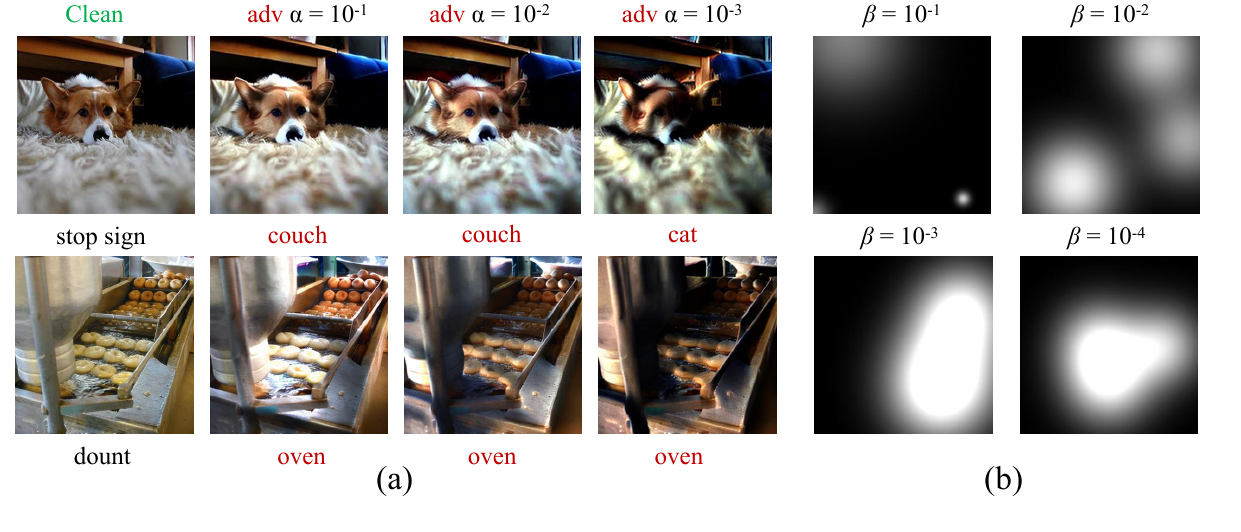}
  \caption{\textbf{Ablation Study Results on Weight Factors.} (a) Visual examples of illumination changes as \(\alpha\) varies, along with the corresponding predicted labels in OpenCLIP. (b) Visualization of the final optimized illumination as \(\beta\) varies across the same sample with three light sources.}
  \label{fig:xiaorong3}
\end{figure}

\noindent\textbf{Weight Factors.} We analyze the effectiveness of the \(\alpha\) and \(\beta\) parameters by visualizing the results for different values in Fig.~\ref{fig:xiaorong3}. The results show that when \(\alpha = 0.1\) and \(\beta = 0.01\), we achieve both high naturalness and attack effectiveness simultaneously.

\section{Conclusion}
\label{sec:Conclusion}

In this paper, we propose \textbf{ITA}, a novel framework designed to generate realistic illumination-aware adversarial examples. By innovatively modeling global illumination using parameterized point light sources and integrating physics-based lighting reconstruction, ITA provides fine-grained control over lighting conditions, enabling diverse and realistic illumination variations. Additionally, ITA incorporates perceptual and diversity constraints, ensuring the generated illumination-aware adversarial examples are both natural and effective. Extensive experiments demonstrate that ITA significantly degrades the performance of advanced VLMs, revealing previously unrecognized vulnerabilities to illumination shifts and emphasizing the urgent need to enhance the illumination-aware robustness of VLMs.

{
    \small
    \bibliographystyle{ieeenat_fullname}
    \bibliography{main}
}

\end{document}


\maketitlesupplementary
\section*{Overview}
This supplementary material provides essential details that complement our main paper. Sec.~\ref{sec:a} presents the ITA update formulation, which builds upon the canonical form of CMA-ES~\cite{hansen2016cma,golovin2017google}. Sec.~\ref{sec:b} lists the specific class names of the 30 selected categories in the COCO dataset. Sec.~\ref{sec:c} provides all prompt templates given to GPT-4o and GPT-4 for the Illumination Natural Score and consistency and correctness metrics. Sec.~\ref{sec:d} showcases additional visualizations of illumination-aware adversarial examples and their performance across various VLMs.

\section{Details of Optimization Algorithm} \label{sec:a}  
Our method optimizes adversarial illumination distributions through the Covariance Matrix Adaptation Evolution Strategy (CMA-ES)~\cite{hansen2016cma,golovin2017google}. Among various evolutionary optimization algorithms, CMA-ES stands out as one of the most effective approaches, demonstrating superior performance particularly on medium-scale optimization problems (typically involving 3-300 variables)~\cite{golovin2017google}. Its gradient-free nature eliminates the dependency on gradient information, making it an ideal choice for optimizing the adversarial illumination distributions in our framework. Additionally, we employ Learning Rate Adaptation (LRA) and Early Stopping policy for efficient search. These enhancements improve convergence speed and prevent unnecessary iterations, making the approach suitable for real-world applications. The adversarial illumination configuration \(\boldsymbol{\Lambda}\) is parameterized as follows:

\begin{equation}
\boldsymbol{\Lambda} = \mathbf{A} \cdot \tanh(\mathbf{q}) + \mathbf{B}, \quad \text{where} \quad \mathbf{q} \sim \mathcal{N}(\boldsymbol{\mu}, \boldsymbol{C\Sigma^2}),
\end{equation}
where \(\mathbf{A}\) and \(\mathbf{B}\) scale Gaussian samples into the feasible range, and \(\mathbf{q}\) follows a Gaussian distribution with mean \(\boldsymbol{\mu} \in \mathbb{R}^{4n}\), step-size \(\boldsymbol{\Sigma} \in \mathbb{R}_{>0}\), and covariance matrix \(\boldsymbol{C} \in \mathbb{R}^{4n \times 4n}\). The optimization maximizes the adversarial impact on VLMs:

\begin{equation}
\begin{gathered}
\arg \max_{\boldsymbol{\mu},\boldsymbol{C},\boldsymbol{\Sigma}} \mathbb{E}_{\mathbf{q} \sim \mathcal{N}(\boldsymbol{\mu}, \boldsymbol{C\Sigma^2})} \big[ \mathcal{L}_{\text{Adv}}(X', Y) + \alpha \cdot \mathcal{L}_{\text{Pecp}} + \beta \cdot \mathcal{L}_{\text{Dis}} \big],
\\
\text{where} \quad X' = \mathcal{I}(X, \mathbf{A} \cdot \tanh(\mathbf{q}) + \mathbf{B}).
\end{gathered}
\end{equation} 

The optimization follows three main steps:

\noindent\textbf{1). Sampling:} Generate a population of candidate adversarial illumination parameters from the Gaussian distribution:

\begin{equation}
    \mathbf{q}^{(i)} \sim \mathcal{N}(\boldsymbol{\mu}, \boldsymbol{C\Sigma^2}), \quad i = 1, \dots, K.
\end{equation}

These samples are then transformed into valid illumination configurations:

\begin{equation}
    \boldsymbol{\Lambda}^{(i)} = \mathbf{A} \cdot \tanh(\mathbf{q}^{(i)}) + \mathbf{B}.
\end{equation}

\noindent\textbf{2). Evaluation:} Compute the objective function values for each sample:

\begin{equation}
    f^{(i)} = \mathcal{L}_{\text{Adv}}(X'^{(i)}, Y) + \alpha \cdot \mathcal{L}_{\text{Pecp}} + \beta \cdot \mathcal{L}_{\text{Dis}},
\end{equation}
where \( X'^{(i)} = \mathcal{I}(X, \boldsymbol{\Lambda}^{(i)}) \) represents the relit image under the adversarial illumination conditions, and \(\alpha\) and \(\beta\) denote the weights.

\noindent\textbf{3). Update:} The distribution parameters \(\boldsymbol{\mu}\), \(\boldsymbol{C}\), and \(\boldsymbol{\Sigma}\) are updated using the CMA-ES strategy. In addition, the learning rate adaptation (LRA) and early stopping policy are applied during the update process to enhance convergence and prevent unnecessary iterations.

\begin{equation}
\label{update mu}
    \boldsymbol{\mu} \leftarrow \sum_{i=1}^{K} w_i \mathbf{q}^{(i)},
\end{equation}

\begin{equation}
\label{update C}
    \boldsymbol{C} \leftarrow (1 - c_c) \boldsymbol{C} + c_c \sum_{i=1}^{K} w_i (\mathbf{q}^{(i)} - \boldsymbol{\mu})(\mathbf{q}^{(i)} - \boldsymbol{\mu})^T,
\end{equation}

\begin{equation}
\label{update Sigma}
    \boldsymbol{\Sigma} \leftarrow \boldsymbol{\Sigma} \cdot \exp\left( c_\sigma \left( \frac{\|\mathbf{p}\|}{\mathbb{E}[\|\mathcal{N}(0, I)\|]} - 1 \right) \right),
\end{equation}
where \( w_i \) are the selection weights, \( c_c \) and \( c_\sigma \) are learning rates, and \(\mathbf{p}\) is the evolution path for step-size control. And the \textbf{Learning Rate Adaptation (LRA)} is integrated into the update of \(\boldsymbol{\Sigma}\):

\begin{equation}
\boldsymbol{\Sigma}_{\text{new}} \leftarrow \boldsymbol{\Sigma} \cdot \exp \left( \frac{c_\sigma}{\|\mathbf{p}\|} \right),
\end{equation}
where \(\|\mathbf{p}\|\) is the norm of the evolution path and \(c_\sigma\) is a learning rate controlling the adaptation. Additionally, the \textbf{Early Stopping Policy} is implemented by monitoring the following conditions during the optimization process: 
\begin{equation}
\Delta f_{\text{best}} = |f_{\text{best}}^{(i)} - f_{\text{best}}^{(i-1)}| < \delta,
\end{equation}
where \(f_{\text{best}}^{(i)}\) is the best fitness at iteration \(i\) and \(\delta\) is a small threshold. If the fitness change is below this threshold, early stopping is triggered.
The optimization will stop if the number of iterations exceeds a predefined maximum limit or if there is no improvement in the best fitness value for a specified number of consecutive generations.

\section{Selected COCO Categories}\label{sec:b}
We conduct experiments of zero-shot classification task (Tab.~\textcolor{iccvblue}{1}) on 30 COCO Categories, generating Adv-IT samples from COCO validation set images. The selected categories are enumerated in Tab.~\ref{COCO class}.

\section{Prompt Templates}\label{sec:c}
\begin{figure*}[t]
  \centering
  \includegraphics[height=20cm]{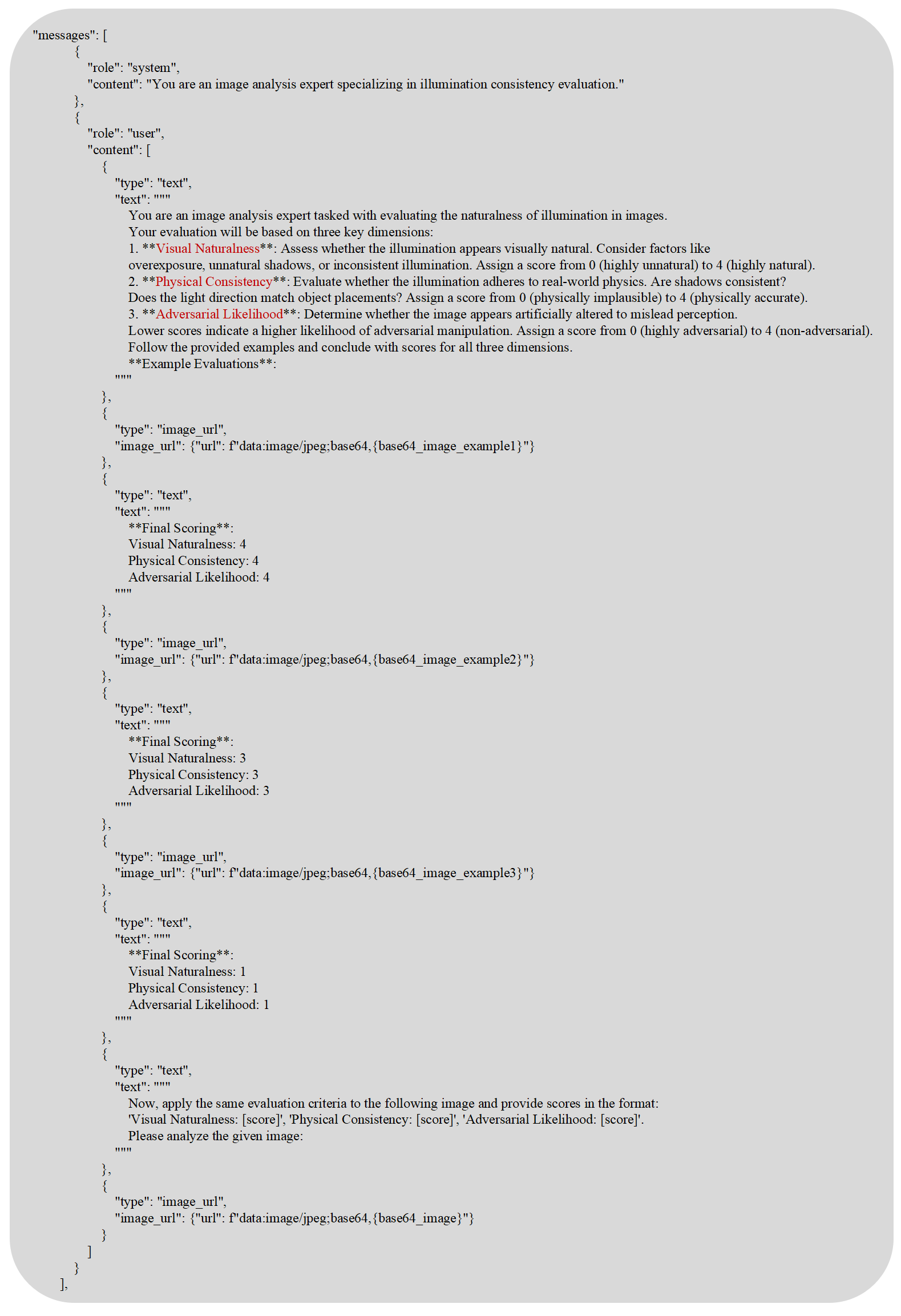}
  \caption{The prompt template for image illumination naturalness evaluation by GPT-4o}
  \label{fig:prompt1}
\end{figure*}

\begin{figure*}[ht]
  \centering
  \includegraphics[height=10.1cm]{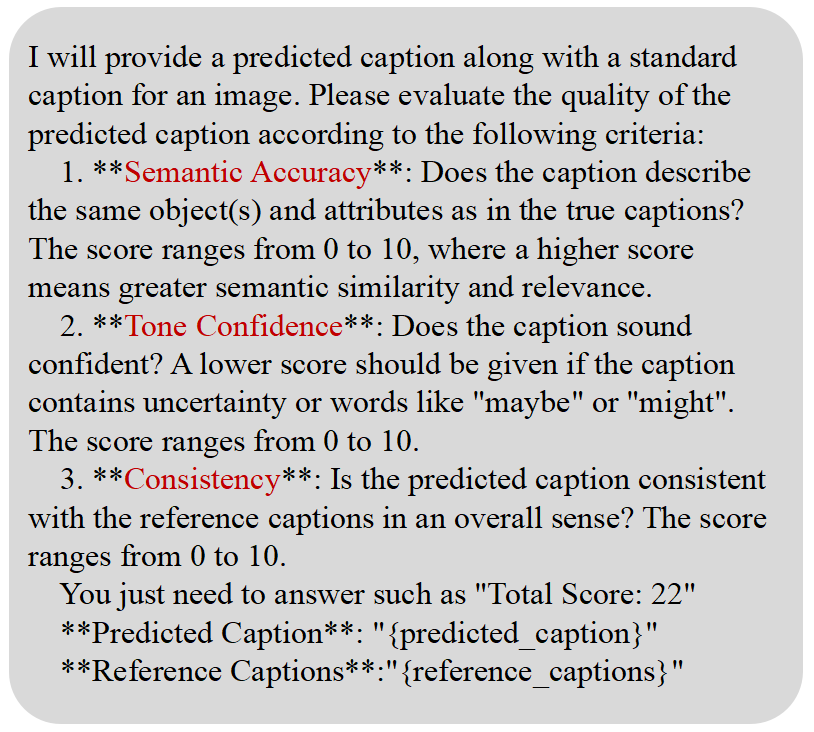}
  \caption{The prompt template for consistency in image captioning tasks.}
  \label{fig:prompt2}
\end{figure*}

\begin{figure*}[ht]
  \centering
  \includegraphics[height=10.1cm]{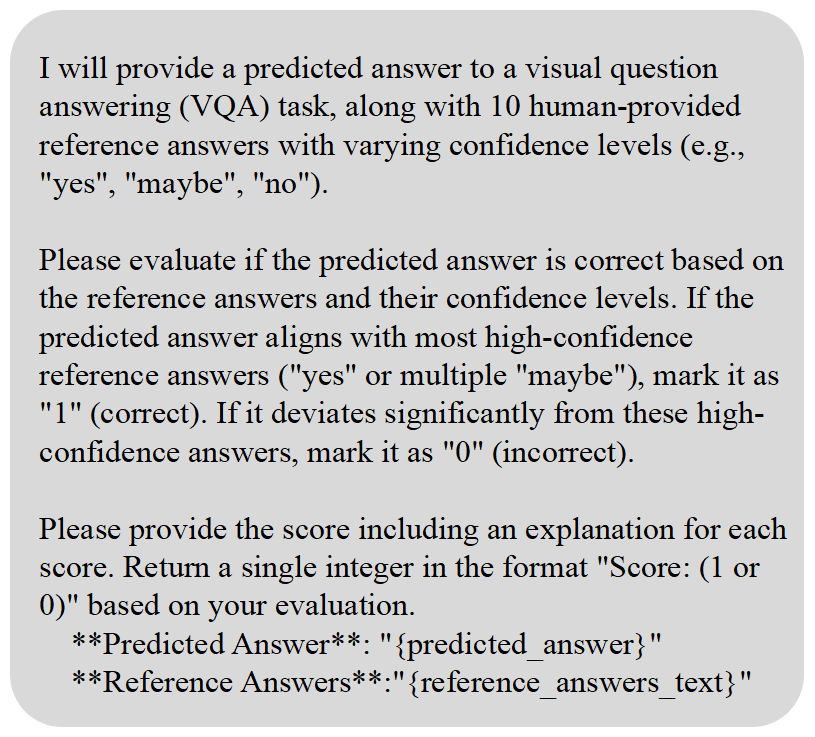}
  \caption{The prompt template for correctness in VQA tasks.}
  \label{fig:prompt3}
\end{figure*}

Fig.~\ref{fig:prompt1} illustrates the prompt template used for evaluating image illumination naturalness in our main experiments (Tab.~\textcolor{iccvblue}{1}). The prompt templates for computing GPT-Score, which are employed to assess image captioning and VQA performance in our main experiments (Tab.~\textcolor{iccvblue}{2} and Tab.~\textcolor{iccvblue}{3}), are presented in Fig.~\ref{fig:prompt2} and Fig.~\ref{fig:prompt3}. Notably, to unify the metrics, these two scores were converted to their respective percentages.

\section{Computational Cost}\label{sec:d}
The computational cost of optimization for each clean sample is approximately 40 GPU minutes. Our experiments on the COCO validation set (300 samples) took 24 GPU hours on 8 NVIDIA RTX 4090 GPUs.

\begin{table}[t]
\caption{The selected 30 categories in COCO dataset.}
\vspace{-0.1cm}
\setlength\tabcolsep{6.0pt}
\renewcommand\arraystretch{1.5}
\centering
\scalebox{0.8}{
\begin{tabular}{c|c|c|c|c|c}
\hline
0  & airplane     & 1  & banana       & 2  & bear         \\ \hline
3  & bed          & 4  & bird         & 5  & boat         \\ \hline
6  & broccoli     & 7  & bus          & 8  & cake         \\ \hline
9  & cell phone   & 10 & clock        & 11 & cow          \\ \hline
12 & dog          & 13 & donut        & 14 & elephant     \\ \hline
15 & fire hydrant & 16 & horse        & 17 & kite         \\ \hline
18 & motorcycle   & 19 & pizza        & 20 & sandwich     \\ \hline
21 & teddy bear   & 22 & traffic light & 23 & stop sign    \\ \hline
24 & toilet       & 25 & train        & 26 & umbrella     \\ \hline
27 & vase         & 28 & zebra        & 29 & sheep        \\ \hline
\end{tabular}}
\label{COCO class}
\end{table}

\section{More Visualization Examples}\label{sec:d}
\begin{figure*}[t]
  \centering
  \includegraphics[width=\textwidth]{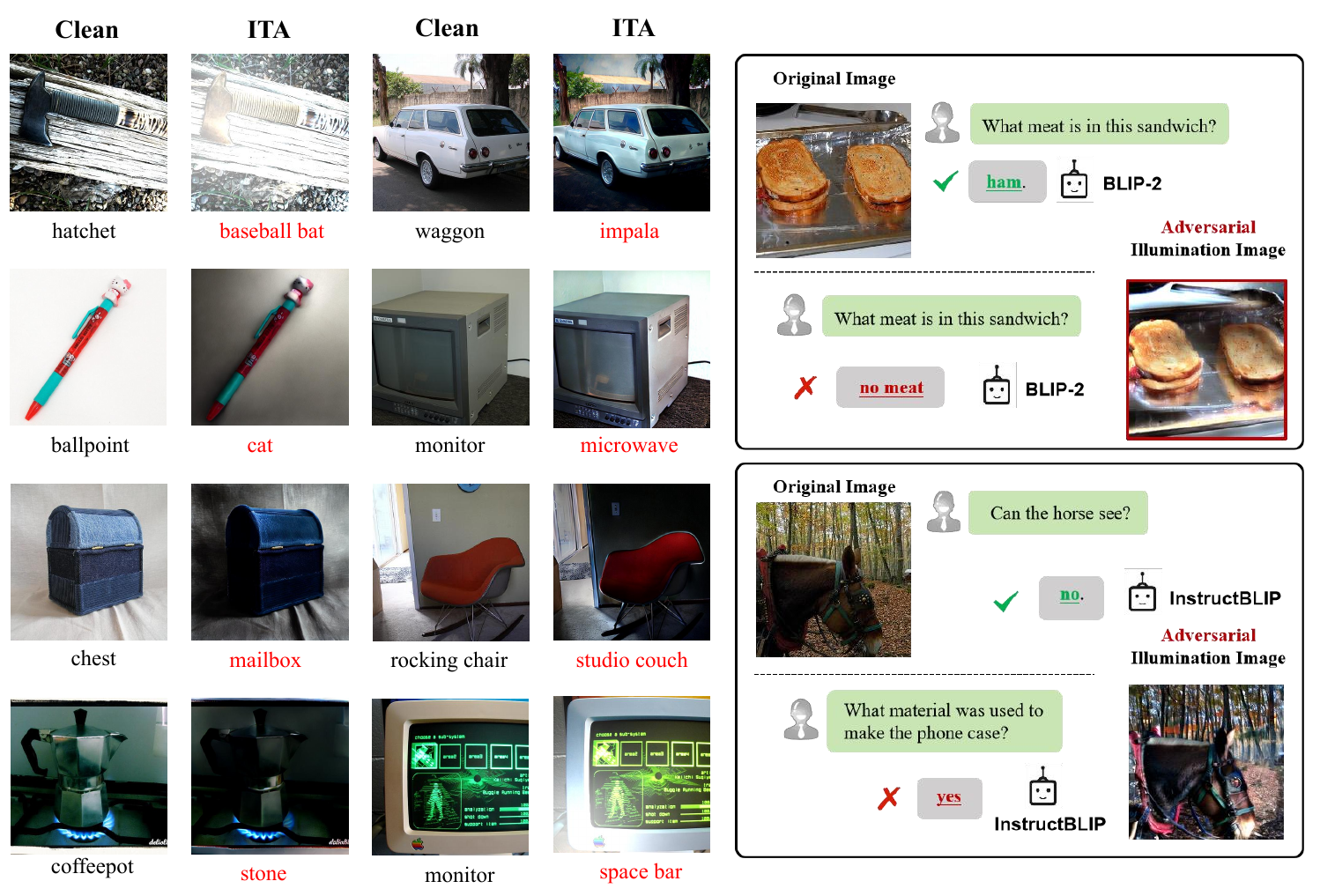}
  \caption{Additional visualization of illumination-aware adversarial examples.}
  \label{fig:vis-3}
\end{figure*}

We provide additional visualization examples of illumination-aware adversarial examples generated by ITA in Fig.~\ref{fig:vis-3}.

\clearpage

\clearpage
{
    \small
    \bibliographystyle{ieeenat_fullname}
    \bibliography{main}
}
